\documentclass{article}

\usepackage{arxiv}
\usepackage{graphicx}
\usepackage[utf8]{inputenc} 
\usepackage[T1]{fontenc}    
\usepackage{hyperref}       
\usepackage{url}            
\usepackage{booktabs}       
\usepackage{amsfonts}       
\usepackage{nicefrac}       
\usepackage{microtype}      
\usepackage{lipsum}
\usepackage{setspace}
\usepackage{subcaption}
\title{A Survey on Deep Learning Methods for Semantic Image Segmentation in Real-Time}

\author{
  Georgios Takos \\
  Mountain View, CA \\
  \texttt{georgios.takos@gmail.com} \\
}

\begin{document}
\maketitle

\begin{abstract}
Semantic image segmentation is one of fastest growing areas in computer vision with a variety of applications. In many areas, such as robotics and autonomous vehicles, semantic image segmentation is crucial, since it provides the necessary context for actions to be taken based on a scene understanding at the pixel level. Moreover, the success of medical diagnosis and treatment relies on the extremely accurate understanding of the data under consideration and semantic image segmentation is one of the important tools in many cases. Recent developments in deep learning have provided a host of tools to tackle this problem efficiently and with increased accuracy. This work provides a comprehensive analysis of state-of-the-art deep learning architectures in image segmentation and, more importantly, an extensive list of techniques to achieve fast inference and computational efficiency. The origins of these techniques as well as their strengths and trade-offs are discussed with an in-depth analysis of their impact in the area. The best-performing architectures are summarized with a list of methods used to achieve these state-of-the-art results.

\end{abstract}


\section{Introduction}

Semantic segmentation is one of the fastest growing areas in Computer Vision and Machine Learning. The availability of cameras and over devices have dramatically increased the interest in better understanding the context of the scene they are capturing and image segmentation is one of the most important components of this process. When an image is analyzed the following levels of understanding are sought:
\begin{enumerate}
\item Classification, i.e., label the most prominent object of an image \cite{AlexNet}.
\item Classification with localization, i.e., extend the previous solution with a bounding box of the object in question.
\item Object detection, where multiple objects of different types are classified and localized \cite{ObjectDetection}.
\item Semantic segmentation, where every pixel in the image is classified and localized.
\item Instance segmentation, an extension to semantic segmentation where different objects of the same type are treated as distinct objects.
\item Panoptic segmentation, which combines semantic and instance segmentation such that all pixels are assigned a class label and all object instances are uniquely segmented.
\end{enumerate}

\begin{figure}[h]
\centering
  \begin{subfigure}[b]{0.45\textwidth}
    \includegraphics[width=\textwidth]{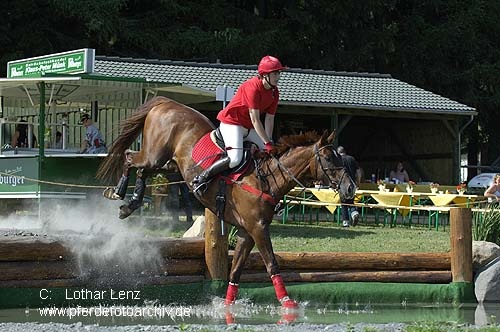}
    \caption{Original image}
    \label{fig:pascal1}
  \end{subfigure}
  \begin{subfigure}[b]{0.45\textwidth}
    \includegraphics[width=\textwidth]{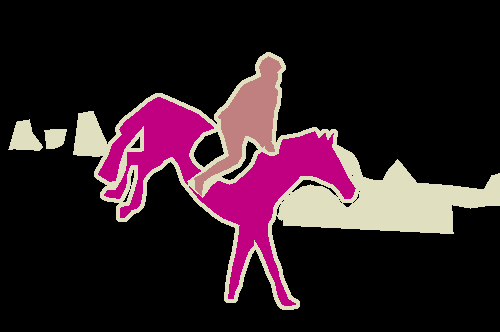}
    \caption{Semantic segmentation ground truth}
    \label{fig:pascal2}
  \end{subfigure}
\caption{PASCAL VOC trainining images}
\end{figure}

The focus of this work is semantic image segmentation, where a pixel-level classification is targeted, and where image pixels  which belong to the same object class are clustered together. An example of this pixel-level classification can be seen in Figure \ref{fig:pascal1} and Figure \ref{fig:pascal2} (see \cite{PascalVOC}). The original image on the left (Figure \ref{fig:pascal1}) can be compared to the semantic segmentation target is the one on the right (Figure \ref{fig:pascal2}), where all objects of interest have been classified.

Semantic segmentation plays an imprortant role in diverse applications such as:
\begin{itemize} 
\item Medical image diagnosis \cite{medical1}, \cite{medical2}.
\item Autonomous driving \cite{SelfDriving1}, \cite{SelfDriving2}.
\item Satellite image processing \cite{Sat1}.
\item Environmental analysis \cite{environment}.
\item Agricultural development \cite{agriculture}.
\item Image search engines \cite{fashion}.
\end{itemize}

In this paper we provide a comprehensive summary of the most recent developments in the area of semantic segmentation with a focus on real-time systems. Efficient techniques for semantic segmentation, where memory requirements and inference time are the main consideration, have not been sufficiently summarized before to the best of the author's knowledge and have been instrumental for the increasing popularity of semantic segmentation across different fields.

This paper is organized as follows: in Section 2 the evolution of traditional image segmentation methods is summarized, followed by, in Section 3, a comprehensive summary of deep learning approaches. Section 4 summarizes the most seminal works on real-time systems, while analyzing the most effective techniques in terms of computational cost and memory load. The following Section enumerates the different datasets that have been used to benchmark different architectures, followed by a Section on the metrics used in the evaluation. The paper concludes with a summary of the performance of different real-time architectures.  

\section{History of Semantic Segmentation} 

One of the earlier approaches to semantic segmentation is thresholding \cite{thresh1}, \cite{thresh2}. It attempts to divide the image to two regions, the target and the background. It works quite well in gray-level images that can be classified in a straightforward manner by using a single threshold. This technique has evolved by taking both local and global threshold values to better capture the image features.

A second technique involves clustering of pixels or regions with similar characteristics, where the image is split into $K$ groups or clusters. All pixels are the assigned a cluster based on a similarity metric that can involve the pixel features (e.g. color, gradient) as well as the relative distance \cite{clust1}. Several popular segmentation techniques have been successfully applied, such as K-means \cite{clust2}, GMMs \cite{clust3}, mean-shift \cite{clust4}, and fuzzy k-means \cite{clust5}.

Edge detection methods \cite{edge1}, have used the fact the edges frequently represent boundaries that can help in segmenting the image. Different edge types have been used (e.g. step edges, ramp edges, line edges and roof edges). Most popular line edge detection methods include Roberts edge
detection \cite{Lawrence}, Sobel edge detection \cite{Sobel}, and Prewitt edge detection\cite{Prewitt}, which utilize different two-dimensional  masks that when convolved with the image will highlight the edges.

A fourth approach looks at images like graphs, where each pixel is a vertex connected with all other pixels, with the weight of each edge measuring the similarity between the pixels. Similarity measures can use features such as distance, intensity, color, and texture to calculate the edge weights. Image segmentation is then treated like a graph partitioning problem, where graph segments are partitioned based on the similarity of the groups \cite{graph1} -- \cite{graph3}. An affinity matrix is computed and the solution to the graph cut problem is given by the generalized eigenvalue of the matrix.

Conditional Random Fields (CRF), a probabilistic framework that can be used to label and segment data, have been used extensively in image segmentation. In this framework, every pixel (that can belong to any of the target class) is assigned a unary cost, i.e., the price to assign a pixel to a class. In addition, a pair-wise cost is added that can model interactions between pixels. For example, zero cost can be assigned when two neighboring pixels belong to the same class, but non-zero cost when the pixels belong to different classes. The unary costs capture the cost for disregarding a class annotation, while the latter penalize non-smooth regions. The goal of the CRF is to find a configuration where the overall cost is minimized. An excellent explanation of CRFs is in \cite{CRF1}, whereas applications in semantic segmentation can be found in \cite{CRF2}, \cite{CRF3}.  

\section{Deep Learning Approaches to Semantic Image Segmentation}

\subsection {Fully Convolutional Networks}
\begin{figure}[h!]
  \includegraphics[width=\linewidth]{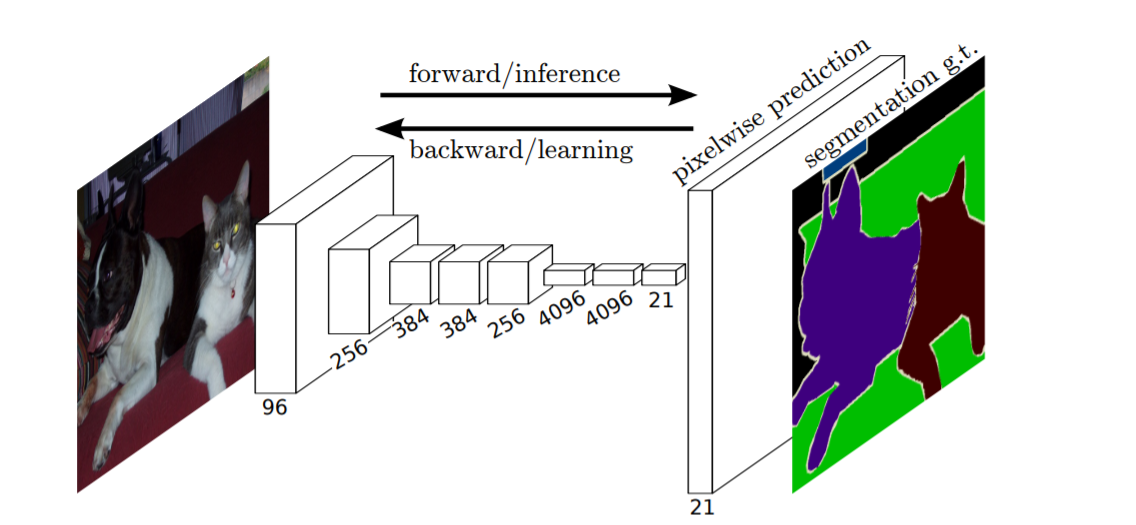}
  \caption{Fully Convolutional Network Architecture (from \cite{FCN}).}
  \label{fig:fcn1}
\end{figure}

Convolution networks  were initially used for classification tasks (AlexNet \cite{AlexNet}, VGG \cite{VGG}, GoogLeNet \cite{GoogLeNet}). These networks first processed the input image with several convolutional layers with increasing number of filters and decreasing resolution, with the last convolutional layer vectorized. The vectorized features were followed by fully connected layers that learn the probability distribution of the classes with a softmax output layer. In FCN \cite{FCN}, the fully connected layers that result in loss of the spatial information were removed from popular architectures (\cite{AlexNet}, \cite{VGG}, and \cite{GoogLeNet}) and replaced with a layer that allow the classification of the image on a per-pixel basis (see Figure \ref{fig:fcn1}). The replacement of fully-connected layers with convolutional ones had two distinct advantages: (a) it allowed the same network architecture to be applied to an image of any resolution and, (b) convolutional layers have fewer parameters which allowed for faster training and inference. This novel approach resulted in state-of-the-art results in several image segmentation milestones and is assumed one of the most influential in the area.

\subsection {Encoder-Decoder Architecture}

\begin{figure}[h!]
  \includegraphics[width=\linewidth]{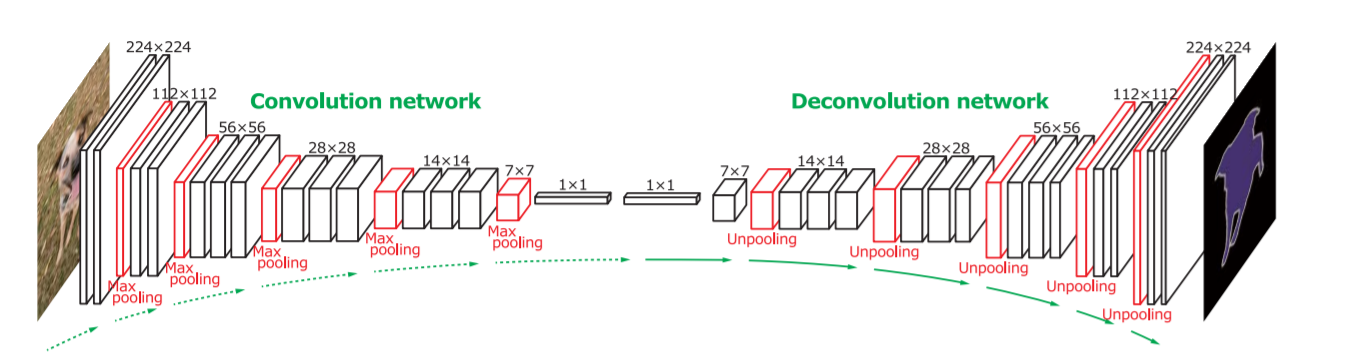}
  \caption{DeconvNet Architecture (from \cite{DeconvNet}).}
  \label{fig:deconvnet1}
\end{figure}

In DeconvNet \cite{DeconvNet}, the authors noted that the approach in \cite{FCN} was leading to loss of information due to the the absence of real deconvolution and the small size of the feature map. They then proposed the architecture in Figure \ref{fig:deconvnet1}, where a multi-layer deconvolution network is learned. The trained network is applied to individual object proposals using fully-connected CRF to obtain instance-wise segmentations, which are combined for the final semantic segmentation. The encoder architecture is based on \cite{VGG}.

\begin{figure}[h!]
  \includegraphics[width=\linewidth]{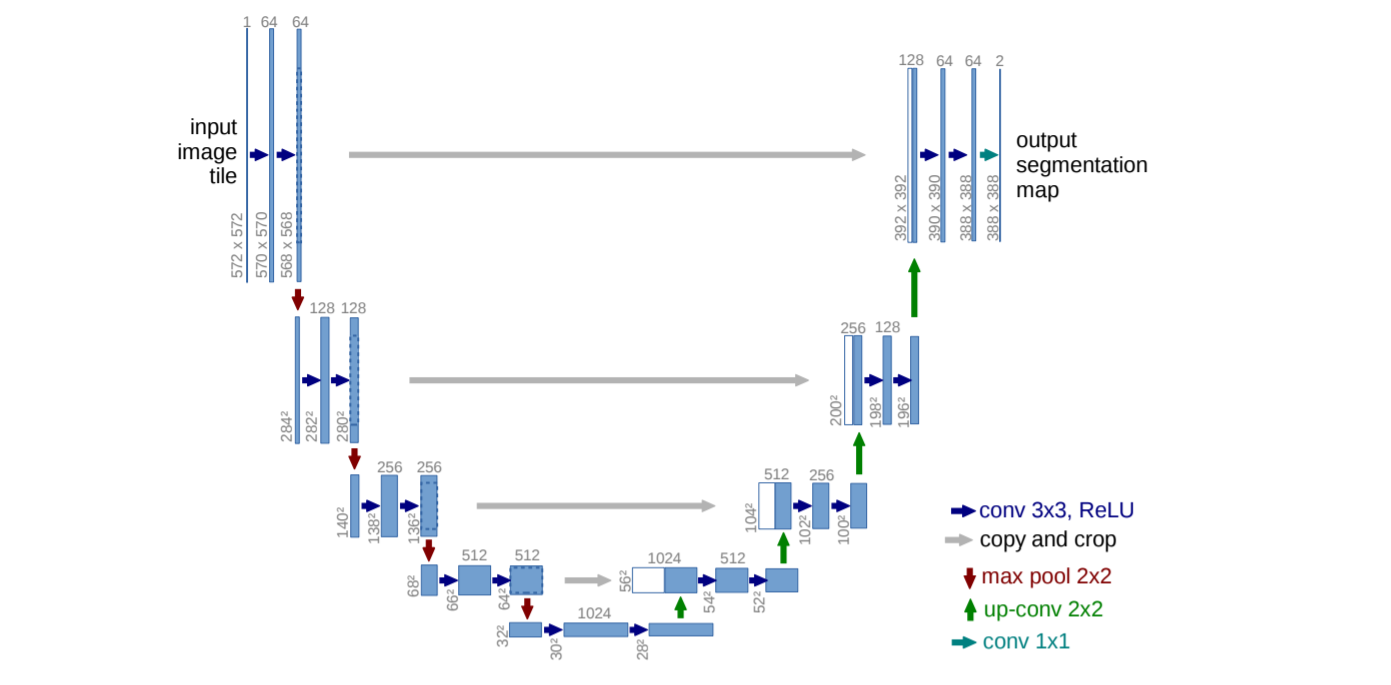}
  \caption{UNet Architecture (from \cite{Unet}).}
  \label{fig:unet1}
\end{figure}

In parallel to \cite{DeconvNet}, decoder/encoder architectures were also used for medical applications \cite{Unet}. The authors proposed an architecture that works well when little training data is available (30 images), which, with appropriate data augmentation can lead to state-of-the-art performance. In Figure  \ref{fig:unet1} the decoder part on the left (contracting path according to the authors) downsamples the image while increasing the number of features. On the upsampling path, the opposite procedure is followed (i.e., increasing the image resolution while decreasing the number of features), while concatenating the corresponding encoder layer. They also proposed a weighted loss around different regions in order to achieve more accurate class separation.

A similar architecture to \cite{Unet} was proposed in SegNet \cite{SegNet}, where the authors used VGG \cite{VGG} as the backbone encoder, removed the fully connected layers, and added a symmetric decoder structure. The main difference is that every decoder layer uses the max pooling indices from the corresponding encoder layer, as opposed to concatenating it. Reusing max-pooling indices in the decoding process has several practical advantages: (i) it improves boundary delineation , (ii) it reduces the number of parameters enabling end-to-end training, and (iii) this form of upsampling can be incorporated into any encoder-decoder
architecture. Although it was originally published in 2015, it originally received little traction until 2017, and has since become one of the most referenced works in semantic segmentation.

\subsection{Conditional Random Fields with Neural Networks}

Conditional Random Fields (CRFs), were one of the most popular methods in semantic segmentation before the arrival of deep learning. CRFs, however, due to their slow training and inference speeds, as well as the difficulty to learn their internal parameters, lost part of their appeal. On the other hand, CNNs by design are not expected to perform well in boundary regions, where two or more classes intersect, or can lose high-level information through the multiple processing stages.  

\begin{figure}[h!]
  \includegraphics[width=\linewidth]{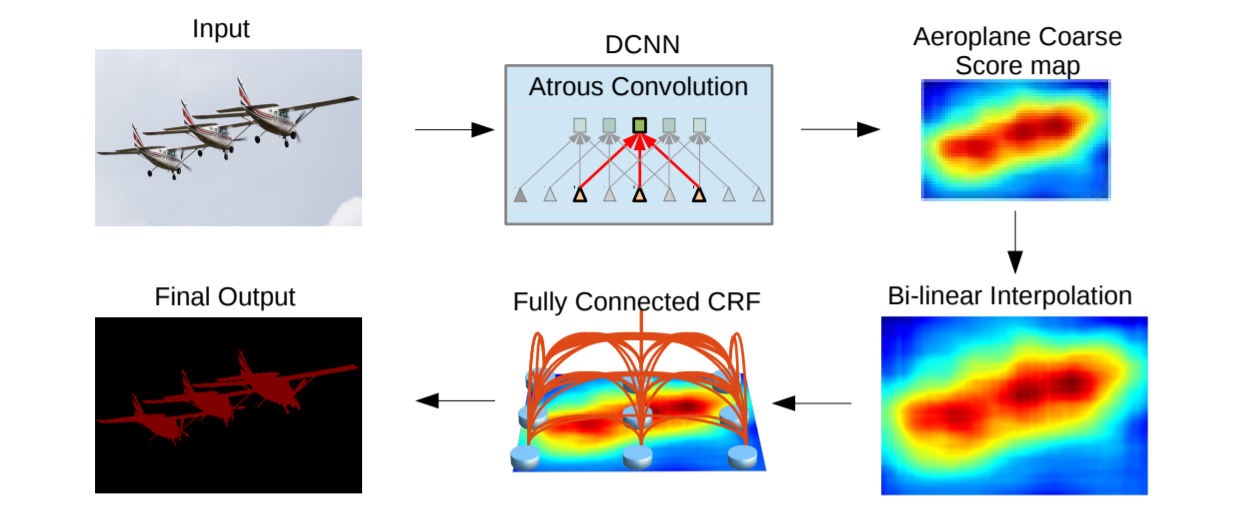}
  \caption{DeepLab Architecture (from \cite{CRF_DL2}).}
  \label{fig:DeepLab}
\end{figure}

The authors of \cite{CRF_DL1} pooled the two approaches by combining the responses at the final neural network layer with a fully connected Conditional Random Field. This way, the model’s ability to capture fine details is enhanced by incorporating the local interactions between neighboring pixels and edges. This work evolved into DeepLab \cite{CRF_DL2}, where several improvements were added (e.g., atrous spatial pyramid pooling), and several variants were proposed. The basic idea can be best explained by Figure \ref{fig:DeepLab}: a fully convolutional network is used to get a coarse score map for the different classes. The image is then upsampled to it full resolution and the CRF is then deployed to better capture the object boundaries. DeepLab achieved state-of-the-art performance in multiple segmentation datasets with an inference time of 125ms or 8 frames per second (FPS).

In the previous work, CRFs are not trained jointly with the fully convolutional network. This can lead to suboptimal end-to-end performance. In \cite{CRF_RNN}, the authors proposed to formulate CRF as an RNN to obtain a deep network that has desirable properties of both CNNs and CRFs. The two networks are then fully integrated and trained jointly to achieve top results on the PASCAL VOC 2012 segmentation benchmark \cite{PascalVOC}.

\subsection{Feature Fusion}

Semantic segmentation involves the task of classifying an image on a pixel level. A lot of the techniques in the area have focused on getting the details of the image right, whereas the context at different stages gets lost. The authors in \cite{FeatureFusion1} suggest enhancing the performance of fully convolutional networks by adding global context to help clarify local confusions. In particular, they propose using the average feature for each layer to augment the features at each location, and thus use the combined feature map to perform segmentation. The effect of global context can be significant; a lot of the misclassified pixels using traditional fully convolutional networks can be recovered when the global context clarifies local confusion and, as a result, a smoother segmentation output is produced.

\begin{figure}[th!]
  \includegraphics[width=\linewidth]{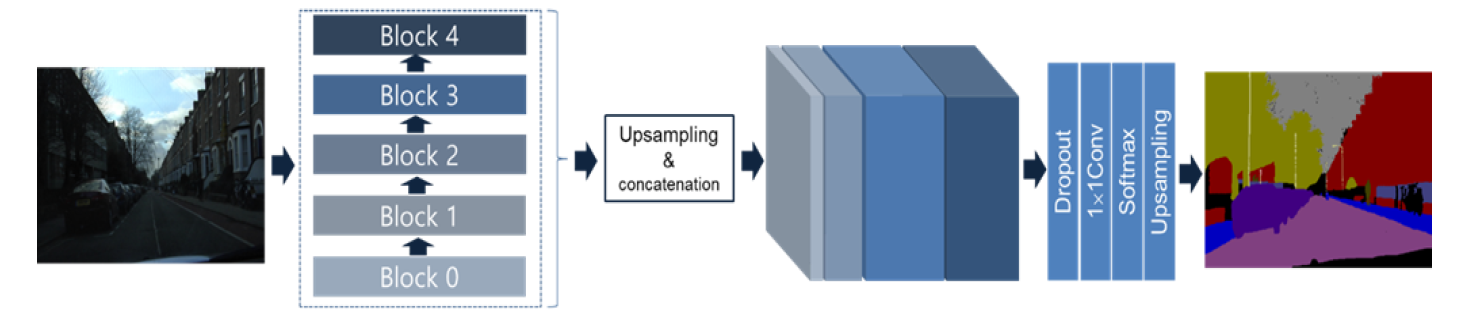}
  \caption{Enhanced Semantic Segmentation Network Architecture (from \cite{FeatureFusion2}).}
  \label{fig:FeatureFusion2}
\end{figure}

The authors in \cite{FeatureFusion2} proposed the Enhanced Semantic Segmentation Network (ESSN), that upsamples and concatenates the residual feature maps from each convolutional layer in order to maintain features from all stages of the network (as seen in Figure \ref{fig:FeatureFusion2}). In \cite{FeatureFusion3}, there is a downsampling stage that extracts feature information, followed by an upsampling part to recover the spatial resolution. The features of the corresponding pooling and unpooling layers are upsampled and concatenated, before the final prediction stage that produces the segmentation output. This fusion at multiple levels of the features maps was evaluated on three major semantic segmentation datasets and achieved promising results.

\subsection {Generative Adversarial Networks}

Generative Adversarial Networks (GANs) were originally introduced in \cite{GAN} as a generative model for unsupervised learning, where the model learns to generate new data with the same statistics as the training set. Its first demonstration was on images, where the artificially generated images looked very similar to those of the training set. Since then, GANs have had quite an impact in diverse areas such as astronomical images \cite{GAN0}, 3D object reconstruction \cite{GAN1}, and image super-resolution \cite{GAN2}.

The idea to apply GANs in semantic segmentation was first introduced in \cite{GAN3}, where the authors used two different networks. First, a segmentation network that took the image as an input and generated per-pixel predictions much like the traditional CNN approaches described earlier in this work, and, second, an adversarial network that discriminates segmentation maps coming either from the ground truth or from the segmentation network. The adversarial network takes as input the image, the segmentation ground truth, and the segmentation network output and outputs a class label (1 for ground truth and 0 for synthetic). An adversarial term is added to the cross-entropy loss function. The adversarial term encourages the segmentation model to produce label maps that cannot be distinguished from ground-truth ones and lead to improved labeling accuracy in the Stanford Background and PASCAL VOC 2012 datasets.

\begin{figure}[h!]
  \includegraphics[width=\linewidth]{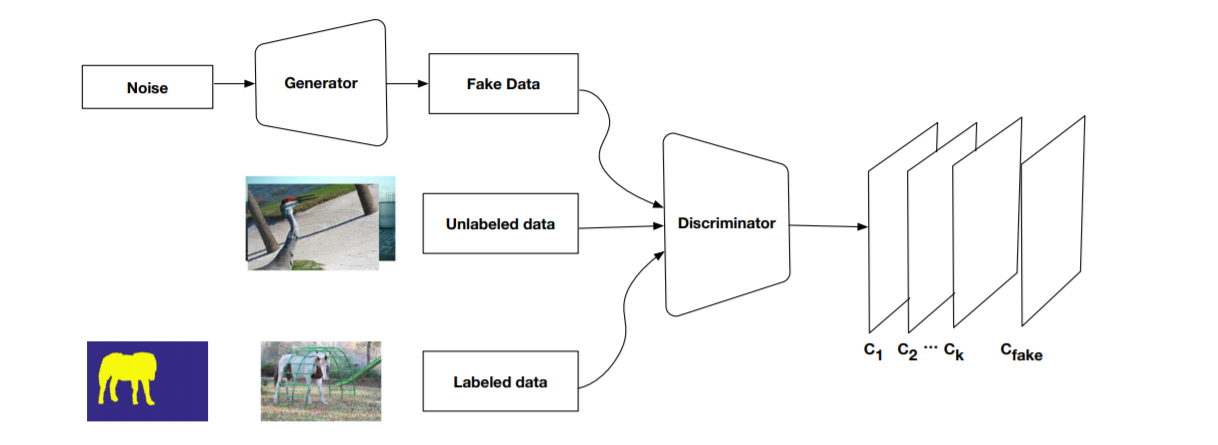}
  \caption{Semi-Supervised Convolutional GAN Architecture (from \cite{GAN4}).}
  \label{fig:GAN1}
\end{figure}

In \cite{GAN4}, the authors proposed a semi-supervised framework – based on Generative Adversarial Networks (GANs) – which consists of a generator network to provide extra training examples to a multi-class classifier, acting as discriminator in the GAN framework, that assigns every sample a label from the K possible classes or marks it as a fake sample (extra class) as seen in Figure \ref{fig:GAN1}. The underlying idea is that adding large fake visual data forces real samples to be close in the feature space, which, in turn, improves multiclass pixel classification. 

The authors of \cite{GAN5}, applied semantic segmentation with GANs in medical images. In a similar fashion to \cite{GAN3}, the adversarial network takes as inputs the original image, the segmentation network output, and the ground truth, and optimize a multi-scale loss function that uses the mean absolute error distance in a min-max fashion. The segmentation network consists of four layers of convolutional stages as in \cite{Unet}, tailored to work with the limited training data sets, and the network significantly outperforms \cite{Unet}.

\subsection {Recurrent Neural Nets (RNNs)}

\begin{figure}[h!]
  \includegraphics[width=\linewidth]{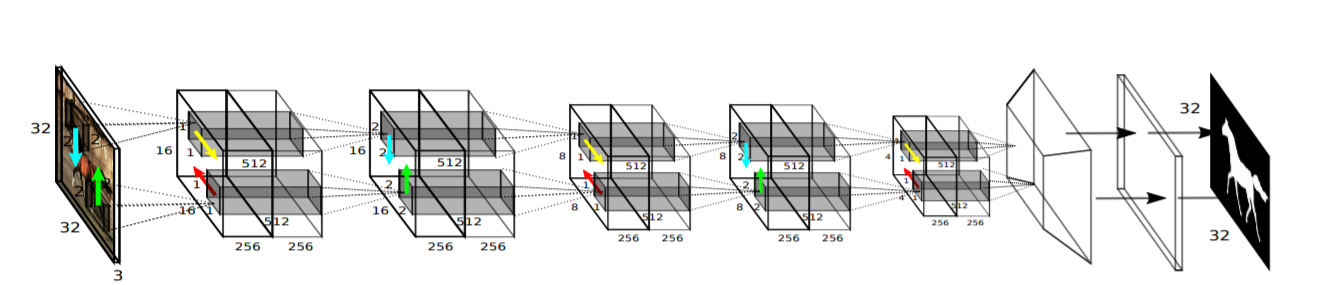}
  \caption{ReSeg Network Architecture (from \cite{RNN1}).}
  \label{fig:RNN1}
\end{figure}

RNNs \cite{RNN} have been widly used for sequential tasks. In \cite{RNN1} the authors proposed ReSeg, which was based on the recently introduced ReNet model for image classification \cite{RNN1_ref}. The latter was tailored to semantic segmentation tasks by transforming each ReNet layer. In particular, each ReNet layer is composed of four RNN (GRUs, see \cite{RNN}) that sweep the image horizontally and vertically in both directions, encoding patches or activations, and providing relevant global information. Moreover, ReNet layers are stacked on top of pre-trained convolutional layers, benefiting from generic local features. Upsampling layers follow ReNet layers to recover the original image resolution in the final predictions. The network architecture can be better understood by looking at Figure  \ref{fig:RNN1}. The first 2 RNNs (blue and green) are applied on small patches of the image, their feature maps are concatenated and fed as input to the next two RNNs (red and yellow) which emit the output of the first ReNet layer. Two similar ReNet layers are stacked, followed by an upsampling layer and a softmax nonlinearity.

Another interesting application to image segmentation was in \cite{RNN2}. There the authors looked at the problem of video segmentation, where consecutive video frames are segmented. One approach would be to independently segment each frame, but this looks like an inefficient approach due to the highly correlated nature of video frames. The authors suggested to incorporate the temporal information by adding an LSTM \cite{RNN}, a type of RNN that can efficiently handle long time dependencies, at different stages in the network and they reported significant performance improvement over their CNN counterparts.

\subsection{Panoptic Segmentation}

Panoptic segmentation \cite{Panoptic}, the task that tries to combine semantic and instance segmentation such that all pixels are assigned a class label and all object instances are uniquely segmented, has shown very promising results \cite{Panoptic1}, \cite{Panoptic2}. The task to provide a coherent scene segmentation incorporating both semantic and instance segmentation seems to be leading to state-of-art results in semantic segmentation over several benchmark data sets as shall be seen later in this report.

\subsection {Attention-based Models}

Attention in deep learning was first introduced in the filed of machine translation \cite{Attention}. The attention mechanism captured long-range dependencies in an effective manner by allowing the model to automatically search for parts of the source sentence that are relevant to predicting a target word.

\begin{figure}[h!]
  \includegraphics[width=\linewidth]{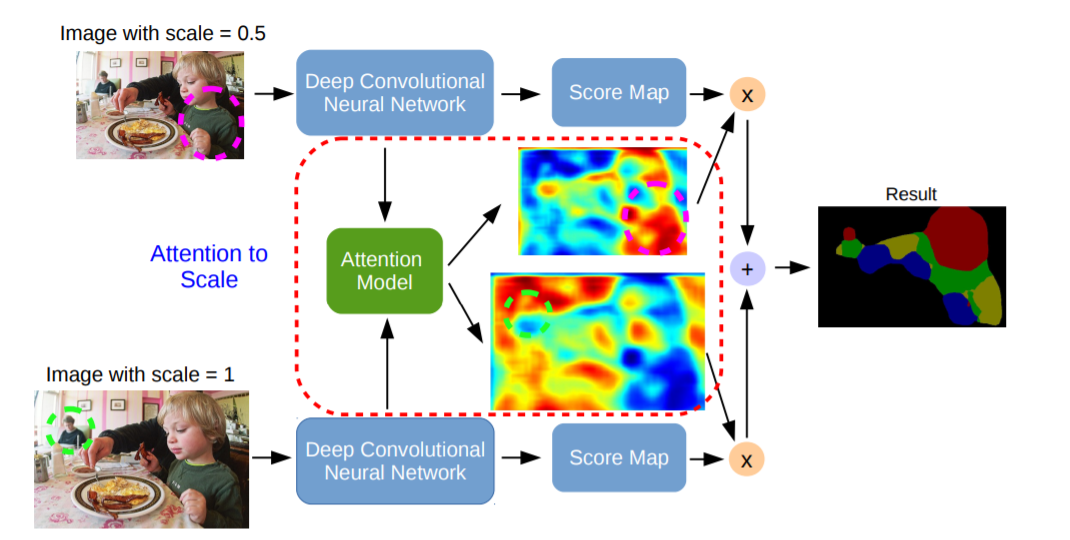}
  \caption{Scale-aware Semantic Image Segmentation Architecture (from \cite{Attention1}).}
  \label{fig:Attention1}
\end{figure}

One interesting way that attention was introduced in semantic segmentation was to incorporate multi-scale features in fully convolutional networks. Instead of the traditional method of feeding multiple resized images to a shared deep network, the authors of \cite{Attention1} proposed an attention mechanism that learns to softly weight the multi-scale features at each pixel location. The convolutional neural network is jointly trained with the attention model as seen in Figure  \ref{fig:Attention1}. As a result, the model learns to scale different size images in an appropriate fashion so that more accurate segmentation is achieved.

 In a similar fashion, \cite{Attention2} tried to address the spatial resolution loss of fully convolutional networks by introducing the feature pyramid attention module. The latter combines the context features from different scales in order to improve classification performance of smaller objects. Attention-aided semantic segmentation networks have been widely used in a variety of applications \cite{Attention3}--\cite{Attention6}.

\section{Real-Time Deep Learning Architectures for Semantic Image Segmentation}

Deep learning based semantic segmentation accuracy has improved significantly from the early approaches. For example, \cite{FCN} achieved 65\% mean intersection over union (mIoU) in the Cityscapes data set \cite{Cityscapes} and 67\% mIoU in the PASCAL VOC 2012 data set \cite{PascalVOCdata}. More recent architectures have outperformed these initial results quite significantly. The authors of HRNet \cite{HRNet}, have build a hierarchical scheme to use images of different scales with an appropriate attention mechanism, like we saw at the end of the previous section. This approach achieves >85\% mIoU in the Cityscapes data set. On the other hand, the authors in \cite{BestPascal}, have used a combination of data augmentation and self-training \cite{SelfTraining} -- that uses noisy labels generated from a model trained on a much smaller labeled data set -- to get >90\% mIoU in the PASCAL VOC 2012 data set.

Computation efficiency, however, is also of paramount importance in several areas like self-driving cars and segmentation on mobile devices, where inference requirements are quite limiting. Computational/memory cost and inference time have to be taken into account when designing a real-time system. In this section, we will go over an exhaustive list of the techniques to build such a system and explain how these improvements were implemented in the literature.

\subsection{Fast Fourier Transform (FFT)}

The well-known convolution theorem \cite{FFT}, states that under suitable conditions the Fourier transform of a convolution of two signals is the pointwise product of their Fourier transforms. The authors in \cite{FFT1} exploited that fact to improve the training and inference time of convolutional networks. A convolution of an image of size $n \times n$ with a kernel of size $k \times k$ will take $\mathcal{O}(n^2*k^2)$ operations using the direct convolution, but the complexity can be reduced to $\mathcal{O}(n^2 \log n)$ by using the FFT-based method. Some additional memory is required to store the feature maps in the Fourier domain, which insignificant compared to the overall memory requirements of a deep neural network.

In \cite{FFT2}, the training and inference algorithms were developed based on FFT and achieved reduced asymptotic complexity of both computation and storage. The authors claim a 1000x reduction in the required number of ASIC cores, as well as a 10x faster inference with a small reduction in accuracy.

\subsection{Pruning}

Storage and memory requirements of a neural network can also be reduced by pruning the redundant weights. In \cite{pruning1}, suggested a three-step approach: first train the network to learn which connections are important, then prune the unimportant connections, and, finally, retrain the network to fine tune the weights of the remaining connections. The number of connections was, therefore, reduced by 9x to 13x with little performance degradation.

The work in \cite{pruning2}, focuses on channel pruning for semantic segmentation networks. They reduce the number of operations by 50\% while only losing 1\% in mIoU using the following stategy: pruning convolutional filters based on both classification and segmentation tasks. This is particularly useful in cases where the network backbone was transferred from an architecture originally built for classification tasks as we have seen earlier in this report. Scaling factors for each convolutional filter are computed based on both tasks and the pruned network is used for inference.

Network pruning is a very active area to improve performance in convolutional neural nets and semantic segmentation, see \cite{pruning3} and \cite{pruning4}, where channel pruning methods can lead to significant compression and speed-up on various architectures that would work on multiple tasks (classification, detection, and segmentation), by either reducing the number of channels on a layer by layer fashion solving a LASSO regression optimization problem, or by pruning the backbone network before transferring it to the segmentation network.

\subsection{Quantization}

Another way to make the network more efficient is by reducing the number of bits required to represent each weight. Typically 32 bits are reserved for the representation of weights. 32-bit operations are slow and have large memory requirements. In \cite{quant1} the authors suggest, among others, to reduce the weight representation to 5 bits, while limiting the number of effective weights by having multiple connections share the same weight, and then fine-tune those shared weights, thus reducing storage requirements.

In Bi-Real Net \cite{quant2}, the authors investigate the enhancement of 1-bit convolutional neural networks, where both the weights and the activations are binary. The performance of these 1-bit CNNs is improved by taking the real-valued output of the batch-normalization layer before the binary activation and connecting it to the real-valued activation of the next block. Thus the representational capability of the proposed model is much higher than that of the original 1-bit CNNs, with only a negligible computational cost.

\subsection{Depthwise Separable Convolutions}

The two previous methods aim at reducing the network size, by either pruning unnecessary components or compressing the weight information. Sifre in his Ph.D. thesis \cite{depth1} introduced a novel method to make 2-dimensional convolutions a lot more computationally efficient called depthwise separable convolution. This idea was picked up by Xception \cite{Xception} and MobileNets \cite{MobileNet} that used slightly modified versions of the original idea to greatly improve the efficiency of their relative architectures. In a regular convolutional layer, the computation complexity depends on (a) the input/output feature map of size $D \times D$ (square feature map is assumed for simplicity), (b) the number of inputs channels $M$, (c) the number of output channels $N$, and (d) the spatial dimension of the kernel $K$. The overall computation requires $D^2 \times K^2 \times M \times N$ multiplications.

In depthwise separable convolutions, the convolution with the filter of size $K \times K \times M \times N$ is broken into two parts. First, a depthwise convolution of a single filter per channel, i.e., of size $K \times K$ for all $M$ input channels, and (b) a pointwise convolution that uses $1 \times 1$ convolutional filters to generate the appropriate output channel dimension. The first operation requires $D^2 \times K^2 \times M$, while the second $D^2 \times M \times N$. The computational improvement is of the order $\max{\big( \mathcal{O}(N), \mathcal{O}(D^2)\big)}$, which can be quite significant especially when the fiter size or depth increases.

\subsection{Dilated Convolutions}

\begin{figure}[h!]
  \includegraphics[width=\linewidth]{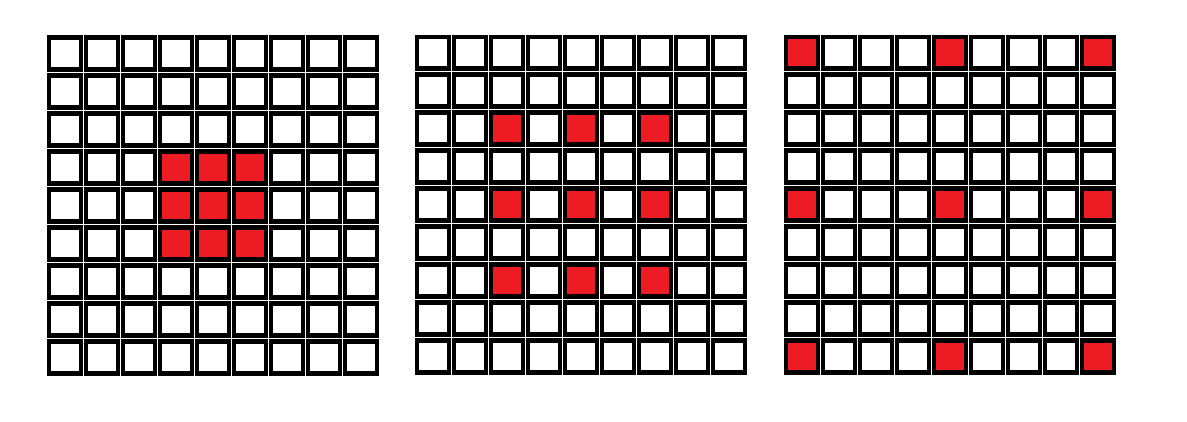}
  \caption{Schematic of $3 \times 3$ dilated convolutional kernel. Left: dilation rate = 1, Center: dilation rate = 2, Right: dilation rate = 4.}
  \label{fig:dilated}
\end{figure}

In their seminal work \cite{dilated1}, the authors introduced dilated convolution to expand the effective receptive field of kernels by inserting zeros between each pixel in the convolutional kernel. As seen on the left side of Figure \ref{fig:dilated}, a $3 \times 3$ kernel will cover nine pixels. However, if a dilation rate of $2$ is introduced, then the eight outer pixels will expand to cover twenty-five pixels, by skipping a pixel (see middle of Figure \ref{fig:dilated}). If the dilation rate further doubles, the coverage will then be of eighty-one pixels as seen on the right of Figure \ref{fig:dilated}. In summary, a kernel of size $K \times K$ with a dilation rate of $N$ will cover $(N-1)*K \times (N-1)*K$ pixels for an expansion of $(N-1) \times (N-1)$. In semantic segmentation tasks, where context is critical for the network accuracy, dilated convolutions can expand the receptive field exponentially without increasing the computation cost. By stacking multiple convolutional layers with different dilation rates, \cite{dilated1} managed to capture image context of increasing receptive fields and was able to significantly improve segmentation performance of previous state-of-the-art works.

In \cite{ESPNet} a new convolutional module was introduced, the efficient spatial pyramid (ESP). ESPNet combines dilated convolution with the depthwise separable convolutions of the previous subsection. In other words, the authors formed a factorized set of convolutions that decompose a standard convolution into a point-wise convolution and a spatial pyramid of dilated convolutions. ESPNet had among the smallest number of parameters of similar works while maintaining the largest effective receptive field. This work is especially interesting since it introduced several new system-level metrics that help to analyze the performance of CNNs.

\subsection{Width and Resolution Multipliers}

In \cite{MobileNet}, the authors explored several ways to further reduce the network complexity. They introduced two hyperpameters: (a) the width multiplier that would produce thinner models and (b) the resolution multiplier that would reduce input resolution. In the former case, the authors scaled down the computational requirements of every layer in a uniform fashion by scaling the number of input and out channels by a factor $\alpha$. From the analysis of the depthwise separable convolutions, it can be seen that the original computational complexity of $D^2 \times K^2 \times M + D^2 \times M \times N$ becomes  $D^2 \times K^2 \times M \times \alpha + D^2 \times M \times N \times \alpha^2$, for an overall reduction of somewhere between $\alpha$ and $\alpha^2$. The basic idea is to find an appropriate scaling factor to define a new smaller model with a reasonable accuracy, latency, and size trade off.

On the other hand, the resolution multiplier $\rho$ can scale the input image dimensions by a factor of $\rho^2$ leading to an overall computational cost of $D^2 \times K^2 \times M \times \rho^2 + D^2 \times M \times N \times \rho^2$. Again this reduction process should be optimized with the accuracy, latency, and size in mind. The two methods can be combined for further improvements.

\subsection{Early Downsampling}

A similar idea was presented in \cite{ENet}, where a number of design choices were discussed based on the authors experimental results and intuitions. In particular, very large input frames are very expensive computationally and it is a good idea to downsample these frames in the early stages of the network, while keeping the number of features relatively low. This downsampling does not have a severe impact in the overall performance because the visual information is typically highly redundant and can be compressed into a more efficient representation. Furthermore, it is noted that the first few layers do not really contribute to the classification task but rather provide useful representations for the subsequent layers. On the other hand, filters operating on downsampled images have a larger receptive field and can provide more context to the segmentation task.

Since the downsampling can lead to loss of spatial information like exact edge shape, ENet follows the paradigm set in SegNet \cite{SegNet}, where the indices of elements chosen in max pooling layers are stored and used to produce sparse upsampled maps in the decoder, therefore, partially recovering spatial information, with small memory requirements.

\subsection{Smaller Decoder Size}

Another design choice discussed in \cite{ENet} was the fact that, in the typical encoder/decoder structure of a semantic segmentation network, the two subcomponents do not have to be symmetric. Encoders need to be deep in order to capture the features in a similar fashion to their classification counterparts. Decoders, however, have one main task: to upsample the compressed feature space in order to provide pixel-level classification. The latter can be achieved with a much less deep architecure, providing significant computational savings.

\subsection{Efficient Grid Size Reduction}

The authors in \cite{Inception} noticed that because the pooling operation can lead to a representational bottleneck, it is typically compensated by increasing the number of channels used before the pooling operation. Unfortunately, this means that the doubling of the filters is effectively dominating the computational cost. Reversing the order of the convolution/pooling operations would definitely improve the computation speed, but would not help with the representational bottleneck. What the authors suggest is to perform pooling operation in parallel with a convolution of stride 2, and concatenate the resulting filter banks. This technique allowed the authors of \cite{ENet} to speed up inference time of the initial block by a factor of 10.

\subsection{Drop Bias Terms}

Bias terms do not have significant impact on the overall performance of a semantic segmentation network and are typically dropped.

\subsection{Stack Multiple Layers with Small Kernels}

Total computational cost increases with the square of the kernel size. In \cite{VGG}, it was argued that having multiple convolutional layers with small kernel size is superior to having a single layer with a larger kernel for two reasons: (a) by stacking three $3 \times 3$ convolutional layers correspond to the same effective receptive field of a $7 \times 7$ layer while reducing the number of parameters to almost half and, (b) by incorporating three non-linear rectification layers instead of a single one, the decision function is made more discriminative.

\subsection{Channel Shuffle Operation}

\begin{figure}[h!]
  \includegraphics[width=\linewidth]{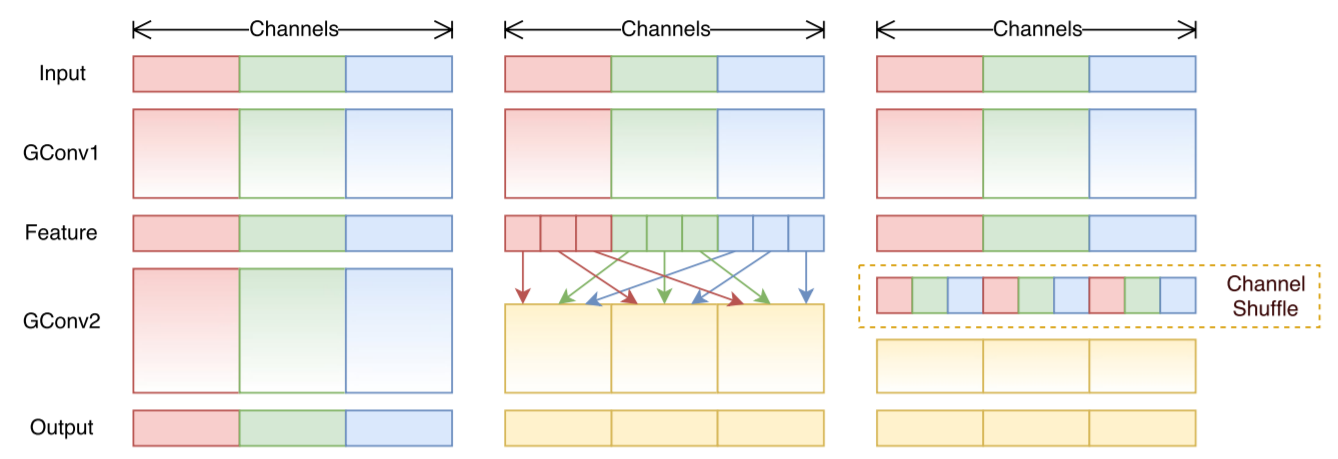}
  \caption{Channel Shuffle Architecture (from \cite{ShuffleNet}).}
  \label{fig:shuffle}
\end{figure}

Grouped convolutions were first introduced in \cite{AlexNet} to distribute the model over multiple GPUs. It uses multiple convolutions in parallel in order to derive multiple channel outputs per layer. \cite{ResNext} showed that the use of grouped convolutions can improve accuracy in classification tasks. However, this architecture become less efficient when applied to much smaller networks, where the performance bottleneck is the large number of dense $1 \times 1$ convolutions. The authors in \cite{ShuffleNet} propose a novel channel shuffle operation to overcome this difficulty. In particular, on the left side of Figure \ref{fig:shuffle} a typical group convolution can be seen with two stacked layers of convolutions and an equal number of groups. If a group convolution is allowed to obtain data from different groups then the input and output channels are fully related (see middle portion of Figure \ref{fig:shuffle}). However, the operation above can be efficiently implemented by following the process at the right of Figure \ref{fig:shuffle}. By adding a channel shuffle operation the output channel dimension is reshaped, transposed and the flattened before being fed to the subsequent layer. Channel shuffle reduces the number of operations by a factor of $g$, which is the number of groups.

\subsection{Two Branch Networks}

Downsampling the original image can lead to significant improvements in the inference time of a semantic segmentation architecture, but it can lead to significant loss of spatial detail. Two branch networks try to reconcile this by introducing two separate branches: (a) a relatively shallow branch that uses the full resolution image that captures the spatial details, and (b) a deeper branch with a downsampled image that will be able to learn efficiently the features for an effective classification outcome. The two branches can share layers as to further improve computational complexity (see \cite{FastSCNN}) or can have different backbone architectures before being aggregated \cite{BiSeNetV2}. The latter work is composed of a detail branch that uses wide channel and shallow layers to capture the low-level details, a semantic branch with narrow channel s and deep layers to maintain the high-level context, and an aggregation layer to fuse the two kinds of features. As a result, BiSeNet-V2 achieves, arguably, the highest inference speed in semantic segmentation tasks (156 frames per second) while maintaining one of the best mIoU performance.

\subsection{Other Design Choices}

Apart from the computationally efficient methods presented in this section so far, there are a handful of other good design choices that would help maintain good performance despite using a lightweight architecture. For example, a common theme in many of the papers is batch normalization \cite{BatchNorm} that allows for faster and more accurate training process. Also the choice of the activation function can be significant. ReLU is the nonlinearity that most works use in the area, but several researchers have reported improved results with parametric ReLU (PReLU).  Finally, regularization \cite{SpatialDropOut} can help avoid overfitting since in many applications the input image dimension is small compared to the number of parameters in a segmentation deep neural network.

\section{Semantic Segmentation Data Sets}

Several data sets have been generated in order to facilitate faster growth in key areas of semantic segmentation as well as establish performance benchmarks. Table \ref{tab:table} summarizes several of the image sets that have been annotated on a pixel level. It contains diverse datasets that were originally developed for classification tasks, as well as more specialized image sets that are appropriate for specific applications (e.g., self-driving and motion-based segmentation) covering a wide range of scenes and object categories with
pixel-wise annotations. Additional information for each of these datasets will be provided in the remainder of this section.
  
\begin{table}[h]
\centering
\begin{tabular}{|p{3cm}||p{2cm}|p{2cm}|p{2cm}|  }
 \hline
 Data Set & \hfil Images &\hfil Classes & \hfil Year\\
 \hline
COCO &\hfil 164K &\hfil 172 & \hfil2017\\
ADE20K &\hfil 25.2K &\hfil 2693 &\hfil 2017\\
Cityscapes &\hfil 25K &\hfil 30 &\hfil 2016\\
SYNTHIA &\hfil 13K &\hfil 13 & \hfil2016\\
PASCAL Context &\hfil 10.1K &\hfil 540 &\hfil 2014\\
SIFT Flow &\hfil 2.7K &\hfil 33 &\hfil 2009\\
CamVid &\hfil 701 &\hfil 32 & \hfil2008\\
KITTI &\hfil 203 &\hfil 13 &\hfil 2012\\
 \hline
\end{tabular}
\caption{Semantic Segmentation Data Set Summary}
\label{tab:table}
\end{table}

\subsection{Common Objects in Context (COCO)} 

Common Objects in Context (COCO) \cite{COCO1} is a large-scale object detection, segmentation, and captioning dataset. It is one of the most extensive datasets available with 330K images of which half are labeled. Semantic classes can be either things (objects with a
well-defined shape, e.g. car, person) or stuff (amorphous background regions, e.g. grass, sky).There are 80 object categories, 91 stuff classes, 1.5 million object instances, and due to the size of the data set it is considered one of the most challenging ones for image segmentation tasks. As a result, the COCO leader board \cite{COCO2} for semantic segmentation consists of only five entries with some of the most seminal works in the area occupying the top spots.
   
COCO-Stuff \cite{COCO3} augments all images of the COCO 2017 data set with pixel-wise annotations for the 91 stuff classes. The original COCO data set already provided outline-level annotation for the 80 thing classes, but COCO-stuff completed the annotation for more complex tasks such as semantic segmentation 

\subsection{PASCAL Visual Object Classes (VOC)} One of the most popular image sets is the PASCAL Visual Object Classes (VOC) \cite{PascalVOC} that can be used for classification, detection, segmentation, action classification, and person layout. The data are available in \cite{PascalVOCdata}, have been annotated, and are periodically updated. For the image segmentation challenge the data include $20$ classes categorized in every-day objects (airplane, bicycle, bird, boat, etc.). The training set consists of 1464 images and the validation set consists of 1449 images. The test set is reserved for evaluation in the PASCAL VOC Challenge, a competition that started in 2005 and it had its most recent data set in 2012. It is a generic data set that includes a variety of scenes/objects and, as a result, it is regularly used to evaluate novel image segmentation approaches. An example of an image and its semantic segmentation can be seen in Figure \ref{fig:pascal1} and Figure \ref{fig:pascal2} respectively.

Several extensions have been made to the former image set, most notably PASCAL Context \cite{PASCALcontext} and PASCAL Part \cite{PASCALpart}. The former annotates the same image with over 500 classes, while the latter breaks down the original objects into several parts and annotates them. Two other PASCAL extensions are: (a) the Semantic Boundaries Data set (SBD) \cite{PASCALSBD}, and (b) the PASCAL Semantic Parts (PASParts) \cite{PASParts}.

\subsection{ADE20K}

\begin{figure}[h]
\centering
  \begin{subfigure}[b]{0.45\textwidth}
    \includegraphics[width=\textwidth]{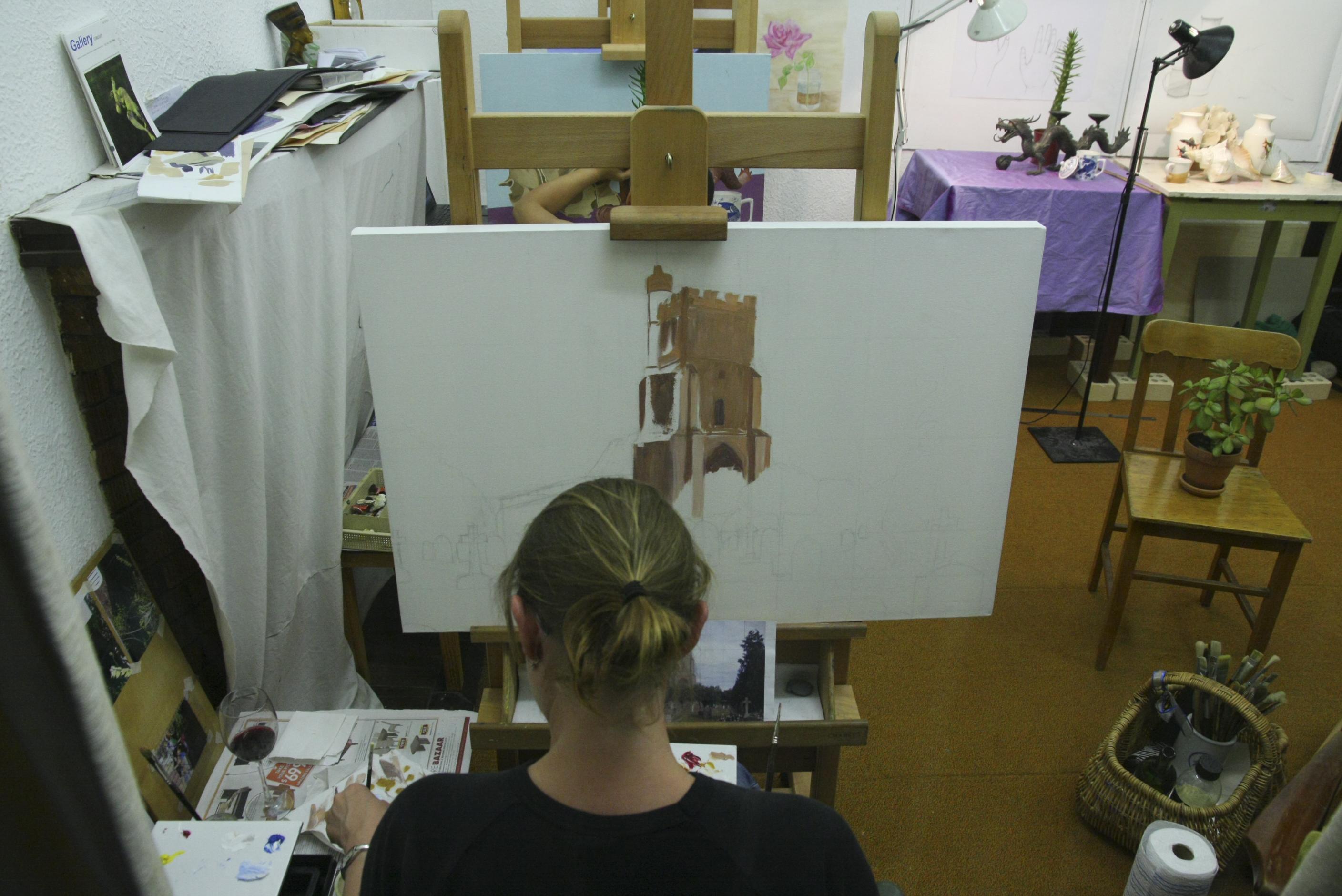}
    \caption{Original image from ADE20K \cite{ADE20K2}}
    \label{fig:ADE1}
  \end{subfigure}
  \begin{subfigure}[b]{0.45\textwidth}
    \includegraphics[width=\textwidth]{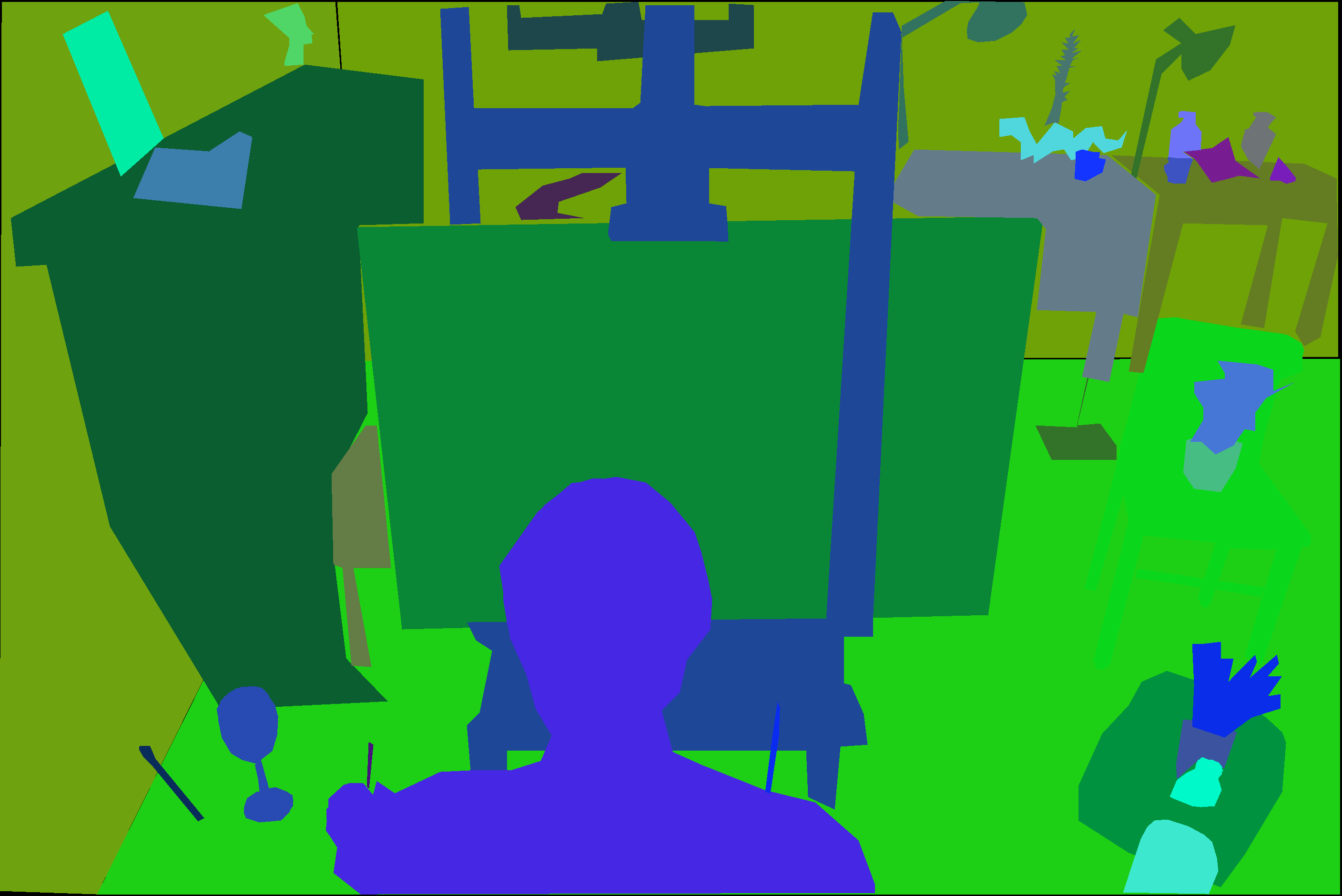}
    \caption{Annotated image from ADE20K \cite{ADE20K2}}
    \label{fig:ADE2}
  \end{subfigure}
\caption{ADE20K trainining images}
\end{figure}

ADE20K \cite{ADE20K1} was developed by the MIT computer vision lab. The authors saw that the datasets available at the time were quite restrictive in the number and type of objects as well as the kinds of scenes. As a result, they collected a data set of 25K images that has densely annotated images (every pixel has a semantic label) with a large and an unrestricted open vocabulary of almost 2700 classes. The images in this data set were manually segmented in great detail, covering a diverse set of scenes, object and object part categories. A single expert annotator, providing extremely detailed and exhaustive image annotations without suffering from annotation inconsistencies common when multiple annotators are used. The detail in the annotation can be seen in Figures \ref{fig:ADE1} and \ref{fig:ADE2}. On average there are 19.5 instances and 10.5 object classes per image. 

For their scene parsing benchmark \cite{ADE20K2}, they selected the top 150 categories ranked by their total pixel ratios and they use the following metrics: (a) pixel accuracy, (b) mean accuracy, (c) mean IoU, and (d) weighted IoU. A little over 20K images were used for the training set, 2K images for the validation, and the remainder was reserved for testing. Stereo video sequences recorded in streets from 50 different cities and annotations involved 30 diverse classes.

\subsection{Cityscapes}

\begin{figure}[h]
\centering
  \begin{subfigure}[b]{0.45\textwidth}
    \includegraphics[width=\textwidth]{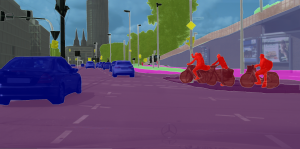}
    \caption{Fine annotation example from Cityscapes \cite{Cityscapes2}}
    \label{fig:Cityscapes1}
  \end{subfigure}
  \begin{subfigure}[b]{0.45\textwidth}
    \includegraphics[width=\textwidth]{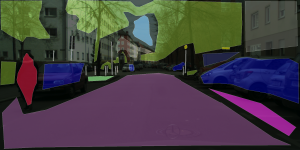}
    \caption{Coarse annotation example from Cityscapes \cite{Cityscapes2}}
    \label{fig:Cityscapes2}
  \end{subfigure}
\caption{Cityscapes trainining images}
\end{figure}

The Cityscapes data set \cite{Cityscapes1} focuses on visual understanding of complex urban street scenes. It has 25K images, 5K of which have high quality pixel-level annotations, while 20K additional images have coarse annotations (i.e., weakly-labeled data) as be seen in Figures \ref{fig:Cityscapes1} and \ref{fig:Cityscapes2}, respectively.
 
Their benchmark suite (found in \cite{Cityscapes2}) involves, among others, a pixel-level semantic labeling task with over 200 entries. It is considered the most diverse and challenging urban scene data sets and, as a result, it is very popular performance evaluation tool.

\subsection{SYNTHIA}

\begin{figure}[h]
  \includegraphics[width=\linewidth]{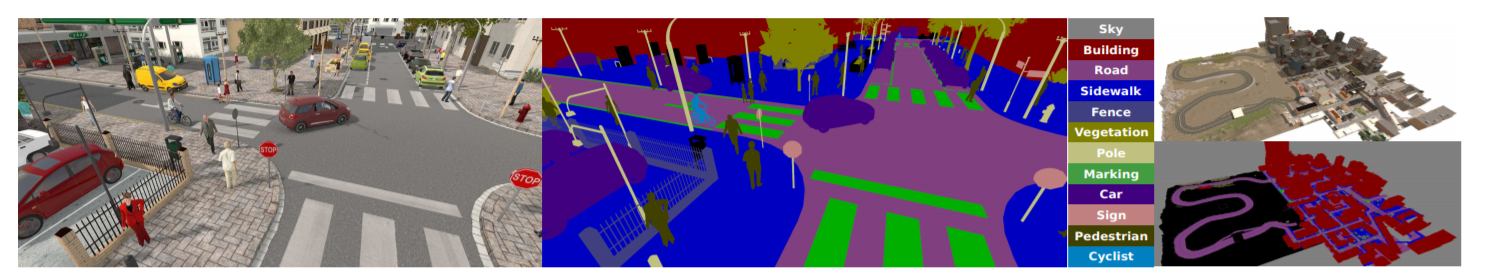}
  \caption{Sample synthetic image from SYNTHIA with its semantic label and a general view of the city (from \cite{SYNTHIA1}).}
  \label{fig:synthia}
\end{figure}

The SYNTHIA data set \cite{SYNTHIA1} is another collection of urban scene images that focuses on self-driving applications. The authors generated realistic synthetic images with pixel-level annotations and tried to address the question of how useful such data can be for semantic segmentation. 13K urban images were created with automatically generated pixel level annotations from 13 categories (e.g. sky, building, road). It was concluded that, when SYNTHIA is used in the training stage together with publicly available real-world urban images, the semantic segmentation task performance significantly improves. An example of a synthetic image from SYNTHIA can be seen in Figure \ref{fig:synthia} as well as the general view of the city used for the image generation. 

\subsection{SIFT Flow}

SIFT Flow \cite{SIFTFlow1} is data set that processed a subset of LabelMe images \cite{SIFTFlow2} in ordered to provide accurate pixel-level annotation of 2688 frames. The top 33 object categories (with the most labeled pixels) were selected mostly from outdoor scenes. The images  were relatively small in size ($256 \times 256$ pixels), and they were generated to evaluate the scene parsing algorithm of the authors.

\subsection{CamVid}

CamVid \cite{CamVid} is another urban scene dataset that includes four high-definition video sequences captured at $960 \times 720$ pixels at 30 frames per second. The total duration of the videos was a little over 22 minutes or around 40K frames. Of the latter, 701 were manually labeled with 32 object classes. Interestingly, the annotation effort took approximately 230 man-hours for an average annotation time of just under 20 minutes. Every annotated image was inspected and confirmed by a second person for accuracy.

\subsection{KITTI}

The last dataset we will be looking at is KITTI \cite{KITTI1} that is quite popular in the self-driving research, since it contains not only camera images, but also laser scans, high-precision GPS measurements and IMU accelerations from a combined GPS/IMU system, sensor data that most autonomous vehicle efforts typically collect. The data were collected while driving in and around Karlsruhe, Germany and it contains over 200 fully annotated images from 13 different classes \cite{KITTI2}. Their semantic segmentation benchmark contains 14 entries, where performance metrics include runtime and environment information to the time-sensitive target applications.

\section{Metrics}

In this section we will be summarizing the basic metrics used to evaluate different semantic segmentation approaches. They either look at the accuracy of the segmentation output (i.e., how close it is to the ground truth) or the efficiency of the approach (i.e., inference time and memory usage).

\subsection{Confusion matrix}

In a segmentation task where there is a total of $C$ classes, the confusion matrix is a $C \times C$ table, where the element in position $(i, j)$ represents the count of pixels that should belong to class $i$ but where classified to belong to class $j$. A good model would result in a confusion matrix that has high counts in its diagonal elements (i.e., correctly classified pixels).

\subsection{Normalized confusion matrix}

It is derived from the confusion matrix, but every entry is normalized by dividing it to the total number of the predicted class $j$. This way all entries are in the range $[0,1]$.

\subsection{Accuracy}

Accuracy, or global accuracy, is the ratio of the correctly classified pixels over the total pixels. It can be derived from the confusion matrix by dividing the sum of the diagonal elements to the total pixels in the image. Accuracy can be misleading especiallly when the classes under consideration are not balanced. For example, if $95\%$ of the pixels are of one class (typically background), a trivial model always predicting this class will lead to $95\%$ accuracy, which does definitely not capture the dependencies of the segmentation task.

\subsection{Mean accuracy}

It is defined as the ratio of correctly classified pixels in each class to total pixels averaged over all classes.  

\subsection{Mean intersection over union}

Mean intersection over union (mIoU) is a metric that addresses the class imbalance weakness of the accuracy metric. In particular, it compares the pixel-wise classification output of a model with the ground truth and finds their intersection and union (i.e., how many pixels were correctly classified as class $i$ for all classes $i$, as well as how many pixels where either classified or were annotated as class $i$ for all classes $i$). The ratio of the intersection over the union (summed over all classes) is the mIoU or Jaccard index. It is robust to class imbalances and is, arguably, the most popular metric when evaluating semantic segmentation tasks.

\subsection{Weighted intersection over union}

This is a small variation of the previous metric to account for the number of pixels per class. It calculates the weighted average of the IoU for each class, weighted by the number of pixels in the class.

\subsection{Precision}

Precision for class $i$ is defined as the proportion of pixels classified as $i$ that were correctly classified. An average precision metric can be defined accordingly for multiple classes. 

\subsection{Recall} 

Recall for class $i$ is defined as the proportion of the actual pixels of class $i$ that were correctly classified. Similarly, an average recall metric can be defined accordingly for multiple classes.

\subsection{F1-score}

F1-score is aggregating the precision/recall metrics by calculating their harmonic mean. It combines features of both and provides information for both types of errors.

\subsection{Frames per second}

All previous metrics measure at the accuracy of the model output, but do not capture the efficiency of the method. One important metric to capture is the inference speed of a network, i.e., the execution time measured in frames per second (fps). It is the inverse of the time to run inference of a new image on a fully trained network. In most real time applications, an fps of 30 or more is required, usually to outperform a typical video frame rate.

\subsection{Memory usage}

Memory usage is a measure of the network size. It can either be measured in number of parameters (for a deep neural network approach), or the memory size to represent the network, or the number of floating point operations (FLOPs) required to run the model.

\section{Performance Summary} 

In this section we will provide summary tables of the best performing models in semantic segmentation. Most papers get evaluated on a subset of the data sets provided earlier in this report and, for most works, computational efficiency is not an critical aspect of the design. As a result, it was decided to summarize the best performing models on the Cityscapes data set \cite{Cityscapes1}, which has been popular with most real-time architectures as an evaluation benchmark. Table \ref{tab:table1} summarizes the top ten performing models with respect to the mIoU with a short summary of the methods used to achieve these results. Anonymous submissions were not included in this section despite occupying some of the top performing spots in the benchmark evaluation. As can be seen in Table \ref{tab:table1}, most entries were published over the past few months, suggesting a very competitive landscape with remarkably fast progress.

\begin{table}[th]
\centering
\begin{tabular}{|p{6cm}||p{1cm}|p{5cm}|p{1cm}|  }
 \hline
 Model & \hfil mIoU &\hfil Methods & \hfil Year\\
 \hline
 \hline
Hierarchical Multi-Scale Attention for Semantic Segmentation \cite{HRNet} &\hfil 85.4 & Hierarchical Attention & \hfil2020\\
 \hline
Naive-Student (iterative semi-supervised learning with Panoptic-DeepLab) \cite{perf1} &\hfil 85.2 & Pseudo-Label Prediction, Data Augmentation & \hfil2020\\
 \hline
Object-Contextual Representations for Semantic Segmentation \cite{perf2} &\hfil 84.5 & Coarse Soft Segmentation, Weighted Aggregation & \hfil2020\\
 \hline
Panoptic-DeepLab  \cite{Panoptic1} &\hfil 84.5 & Panoptic Segmentation & \hfil2020\\
 \hline
EfficientPS: Efficient Panoptic Segmentation  \cite{Panoptic2} &\hfil 84.2 & Panoptic Segmentation & \hfil2020\\
 \hline
Axial-DeepLab  \cite{perf3} &\hfil 84.1 & Panoptic Segmentation, Self Attention & \hfil2020\\
 \hline
Improving Semantic Segmentation via Decoupled Body and Edge Supervision \cite{perf4} &\hfil 83.7 &  Decoupled Multi-scale Feature Training & \hfil2020\\
 \hline
Improving Semantic Segmentation via Video Propagation and Label Relaxation \cite{perf5} &\hfil 83.5 &  Joint Future Frame/Label Propagation, Data Augmentation & \hfil2019\\
 \hline
Hard Pixel Mining for Depth Privileged Semantic Segmentation \cite{perf6} &\hfil 83.4 &  Depth Information, Depth-Aware Loss & \hfil2019\\
 \hline
Global Aggregation then Local Distribution in Fully Convolutional Networks \cite{perf7} &\hfil 83.3 &  Global Aggregation, Local Distribution & \hfil2019\\
 \hline
\end{tabular}
\caption{Cityscapes Pixel-Level Semantic Labeling Task Top Performing Models}
\label{tab:table1}
\end{table}

Table \ref{tab:table2} ranks real-time semantic segmentation works where the performance metric is inference speed (i.e., frames per second (FPS)). Three of the top ten positions are occupied by a single paper \cite{FastSCNN}, which clearly demonstrates the performance/efficiency trade-offs. However, as this table shows, real-time semantic segmentation is a reality and several architectures achieve accuracy close to state-of-the-art semantic segmentation models.

\begin{table}[h]
\centering
\begin{tabular}{|p{5.33cm}||p{1cm}|p{1cm}|p{4.4cm}|p{1cm}|  }
 \hline
 Model & \hfil FPS & \hfil mIoU &\hfil Methods & \hfil Year\\
 \hline
 \hline
FastSCNN (quarter-resolution) \cite{FastSCNN} &\hfil 485 &\hfil 51.9 & Two-branch Networks, Depthwise Separable Convolutions & \hfil2019\\
 \hline
FastSCNN (half-resolution) \cite{FastSCNN} &\hfil 286 &\hfil 63.8 & Two-branch Networks, Depthwise Separable Convolutions & \hfil2019\\
 \hline
FasterSeg \cite{eff1} &\hfil 163 &\hfil 71.5 & Neural Architecture Search & \hfil2020\\
 \hline
LiteSeg \cite{eff2} &\hfil 161 &\hfil 67.8 & Depthwise Separable Convolutions, Dilated Convolutions & \hfil2019\\
 \hline
Partial Order Pruning \cite{eff3} &\hfil 143 &\hfil 71.4 & Neural Architecture Search, Pruning & \hfil2019\\
 \hline
RPNet \cite{eff4} &\hfil 125 &\hfil 68.3 & Early Downsampling, Residual Blocks & \hfil2019\\
\hline
FastSCNN \cite{FastSCNN} &\hfil 123 &\hfil 68 & Two-branch Networks, Depthwise Separable Convolutions & \hfil2019\\
\hline
Spatial Sampling Network for Fast Scene Understanding \cite{eff5} &\hfil 113 &\hfil 68.9 & Smaller Decoder Size & \hfil2019\\
\hline
ESPNet \cite{ESPNet}  &\hfil 112 &\hfil 60.3 & Dilated Convolutions, Depthwise Separable Convolutions & \hfil2018\\
\hline
Efficient Dense Modules of Asymmetric Convolution \cite{eff6} &\hfil 108 &\hfil 67.3 & Dilated Convolutions, Asymmetric Convolutions & \hfil2018\\
\hline
\end{tabular}
\caption{Cityscapes Pixel-Level Semantic Labeling Task Top Performing Real-Time Models}
\label{tab:table2}
\end{table}

\section{Summary}

This work provides an extensive summary of the most recent advances in semantic image segmentation with deep learning methods focusing on real-time applications. It starts with an explanation of the segmentation task and how it differs from similar tasks, continues with a history of early segmentation methods, and provides a detailed description of the different deep learning approaches of the last decade. An extensive list of techniques to improve the efficiency of deep learning networks, by optimizing different aspects of the network, is then provided and the trade-offs in these design choices are explained. The most widely used benchmark data sets are subsequently described, followed by a list of metrics used to evaluate the accuracy and efficiency of the proposed models. Finally, performance tables are provided to summarize the state-of-the-art approaches in the area of semantic segmentation, both from an accuracy perspective, as well as an efficiency one. 

Recent advances in deep learning methods, contemporaneously with a rapid increase in image capturing capabilities, have made image segmentation a crucial tool in a plethora of applications, from medical imaging to time-critical applications such as autonomous driving. This survey summarizes recent breakthroughs that transformed the field of image segmentation and provides a comprehensive insight on the design choices that led to this transformation.

\end{document}